%% file: 00_main.tex
\UseRawInputEncoding
\documentclass{article}
\usepackage[nonatbib, final]{neurips_2022}
\usepackage{hyperref}       
\usepackage{url}            
\usepackage{booktabs}       
\usepackage{amsfonts}       
\usepackage{nicefrac}       
\usepackage{microtype}      
\usepackage[noadjust]{cite}
\usepackage{amsmath,amssymb}
\usepackage{gensymb}
\usepackage{algorithm}
\usepackage[algo2e]{algorithm2e}
\usepackage{color}
\usepackage{amsfonts}
\usepackage{graphicx}
\usepackage{tabulary,multirow,overpic,xcolor}
\usepackage{wrapfig}
\usepackage{sidecap}
\usepackage{wrapfig}
\usepackage{todonotes}
\usepackage{blindtext}
\usepackage{subcaption}
\usepackage{siunitx}
\makeatletter
\let\NAT@parse\undefined
\makeatother
\usepackage[sort&compress, numbers]{natbib}

\hypersetup{
    colorlinks=true,
    linkcolor=blue,
    filecolor=magenta,      
    citecolor=black,
    urlcolor=blue,
}

\newcommand{\cabinet}{CabiNet\xspace}

\newcommand*\samethanks[1][\value{footnote}]{\footnotemark[#1]}

\newcommand{\app}{\raise.17ex\hbox{$\scriptstyle\sim$}}
\newcolumntype{x}[1]{>{\centering\arraybackslash}p{#1pt}}

\newlength\savewidth

\makeatletter\renewcommand\paragraph{\@startsection{paragraph}{4}{\z@}
  {.5em \@plus1ex \@minus.2ex}{-.5em}{\normalfont\normalsize\bfseries}}\makeatother

\title{\bf CabiNet: Scaling Neural Collision Detection for Object Rearrangement with Procedural Scene Generation}

\author{
Adithyavairavan Murali \\
NVIDIA \\
\texttt{admurali@nvidia.com} \\
\And
Arsalan Mousavian \\
NVIDIA \\
\texttt{amousavian@nvidia.com} \\
\And
Clemens Eppner \\
NVIDIA \\
\texttt{ceppner@nvidia.com} \\
\AND
Adam Fishman\thanks{Author is also affiliated with the University of Washington} \\
NVIDIA \\
\texttt{afishman@nvidia.com} \\
\And
Dieter Fox\samethanks \\
NVIDIA \\
\texttt{dieterf@nvidia.com} \\
\vspace{-5mm}
}

\begin{document}
\maketitle


\begin{figure*}[!ht]
  \centering
  \includegraphics[width=0.90 \textwidth]{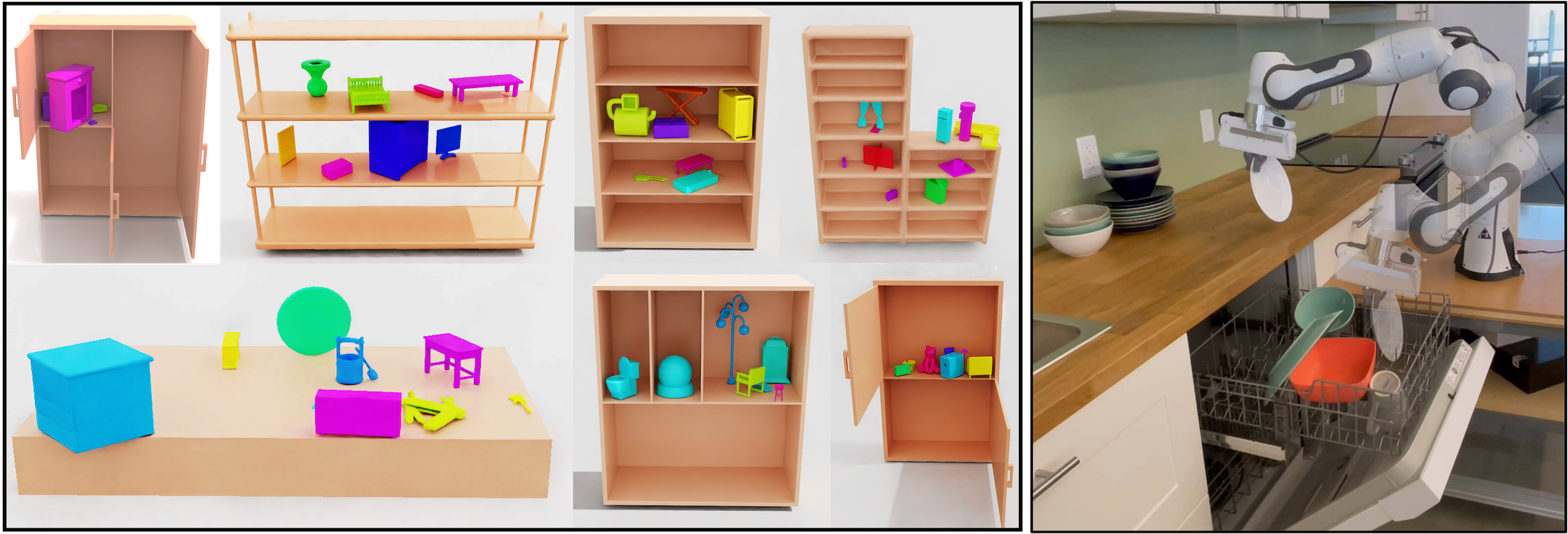}  
  \caption{CabiNet is able to \textit{(right)} perform complex rearrangement tasks in novel, cluttered scenes on the real robot from just partial point cloud observations without object or environment models. The model is trained with over 650K procedurally generated synthetic scenes \textit{(left)}.}
  \label{fig:cover}
\end{figure*}
  
\begin{abstract}
We address the important problem of generalizing robotic rearrangement to clutter without any explicit object models. We first generate over 650K cluttered scenes---orders of magnitude more than prior work---in diverse everyday environments, such as cabinets and shelves. We render synthetic partial point clouds from this data and use it to train our \cabinet model architecture. \cabinet is a collision model that accepts object and scene point clouds, captured from a single-view depth observation, and predicts collisions for SE$(3)$ object poses in the scene. Our representation has a fast inference speed of 7$\mu$s/query with nearly 20$\%$ higher performance than baseline approaches in challenging environments. We use this collision model in conjunction with a Model Predictive Path Integral (MPPI) planner to generate collision-free trajectories for picking and placing in clutter. \cabinet also predicts waypoints, computed from the scene’s signed distance field (SDF), that allows the robot to navigate tight spaces during rearrangement. This improves rearrangement performance by nearly 35$\%$ compared to baselines. We systematically evaluate our approach, procedurally generate simulated experiments, and demonstrate that our approach directly transfers to the real world, despite training exclusively in simulation. Videos of robot experiments in completely unknown scenes are available at: \href{https://cabinet-object-rearrangement.github.io}{cabinet-object-rearrangement.github.io}.

\end{abstract}


\input{01_intro.tex}
\input{02_related_work.tex}

\input{03_method.tex}
\input{04_results.tex}
\input{05_conclusion.tex}
\input{06_acknowledgements.tex}



\clearpage
\bibliographystyle{plain}
\bibliography{bibliography}

\clearpage
\input{appendix}

\end{document}

%% file: 01_intro.tex
\section{Introduction}

Object rearrangement is an important challenge in robotic manipulation and decision making \cite{batra2020rearrangement, garrett2021}. It requires the skills of picking, placing and generating complex collision-free motions in a cluttered environment. The Embodied AI \cite{ehsani2021manipulathor, szot2021habitat, Xiang_2020_SAPIEN, li2021igibson} and Task-And-Motion-Planning (TAMP) \cite{tamp204, garrett2020onlinereplanning} literature have reported impressive results of rearrangement in complex human environments like kitchens. But, they are largely limited to simulation \cite{szot2021habitat, ehsani2021manipulathor, garrett2021}, do not enable unconstrained 6-DOF manipulation \cite{robotsinhomenips2018} or make the strong assumption of state estimation of the environment \cite{Wada:etal:ICRA2022b} and objects \cite{garrett2020onlinereplanning} in the real world. Recent neural rearrangement methods \cite{nerp2021, Murali2020CollisionNet, ifor2022, mahler2017dex, rt12022arxiv} generalize from sensed observations, without requiring state information, and have been demonstrated in the real world. Yet, they are limited to tabletop scenes \cite{nerp2021, Murali2020CollisionNet} or require  expensive data collection on the real robot: 1.5 years on 13 robots in the case of \cite{rt12022arxiv} and six houses in \cite{robotsinhomenips2018}. Overall, there are no rearrangement systems that generalizes to novel challenging environments and works out-of-the-box with minimal engineering effort for each new scene type~\cite{nistenv2021}. The cost of system integration, a major task of which is environment modelling, comes up to 3X of the price of the robot itself \cite{robotiq2021}. In this work, we aim to learn a single representation, trained over 650K scenes in simulation, that enables rearrangement in diverse unknown environments. 

One of the primary challenges in achieving generalization in object rearrangement is the limited availability of rearrangement datasets. Large data has been the driving force behind the success of visual learning~\cite{VisionTransformers2020} and large language models~\cite{CLIP2021}. There have been some recent attempts at large-scale learning in simulation, such as the Habitat~\cite{szot2021habitat} and ManipulaTHOR~\cite{ehsani2021manipulathor} efforts. However, these actions are abstract and are not realistic. For example, objects simply stick to the gripper based on proximity, thereby sidestepping the complex dynamics of pick-and-place that are explicitly addressed in prior work in the robotic grasping literature \cite{mahler2018binpicking,6dofgraspnet, Murali2020CollisionNet, sundermeyer2021contact}. Additionally, the policies learned in these simulated environments are unproven in the real world. To facilitate performance in a physical system, our approach is conditioned on 3D point clouds instead of the RGBD representation commonly used in \cite{ehsani2021manipulathor, szot2021habitat}. Prior work \cite{Murali2020CollisionNet, Danielczuk2021ObjectRU} has demonstrated that point clouds are an effective representation to transfer from simulation-based training to real world observations.

In robotic rearrangement, a fundamental component for planning is collision detection with an unknown environment. Classical TAMP methods typically rely on the complete geometric model of the scene for planning \cite{fcl2012, garrett2021} in the form of a triangular mesh or Signed Distance Field (SDF). 
Visual reconstruction systems such as SLAM \cite{klingensmith2016}, KinectFusion~\cite{KinectFusion2011} and more recently NERF~\cite{Mildenhall2020NeRFRS} are needed to generate a geometric model of the scene. 
Each system has its drawbacks, which may include a long start-up time~\cite{KinectFusion2011, Mildenhall2020NeRFRS}, multi-view requirements ~\cite{Mildenhall2020NeRFRS, klingensmith2016}, costly updates in dynamic scenes~\cite{KinectFusion2011, Mildenhall2020NeRFRS, klingensmith2016}, or poor generalization~\cite{Mildenhall2020NeRFRS}. Instead of explicitly reconstructing the scene for a traditional collision checker, recent neural methods speed up collision detection with learning \cite{faust2021, yip2020} and generalize to partial real-world observations \cite{Murali2020CollisionNet, Danielczuk2021ObjectRU}. \cite{Murali2020CollisionNet} proposed the CollisionNet model to predict collisions from scene point clouds for 6-DOF grasping. \cite{Danielczuk2021ObjectRU} extended this to an architecture that allowed fast collision checking from point clouds, enabling fast sampling-based placing in cluttered tabletop scenes. In this work, we extend this 3D implicit representation to scale to multiple cluttered environments, which we call \cabinet and learn a SDF-based waypoint sampler from this 3D representation. We then apply a Model Predictive Path Integral (MPPI) \cite{mppi2017} algorithm on the GPU to use our \cabinet model to generate  pick-and-place motion trajectories in clutter. In summary, our contributions are as follows:
\begin{itemize}
    \item Scaling up neural collision checking by 30X compared to prior work \cite{Danielczuk2021ObjectRU}, training over nearly 60 billion collision queries. We also learn from over 650K cluttered scenes generated procedurally, which is six orders of magnitude more scene data than prior work on learning rearrangement in simulation \cite{ehsani2021manipulathor}. We train a implicit 3D scene encoder \cabinet from this dataset with over 2.5 million synthetically rendered point clouds.
    \item We demonstrate that \cabinet achieves fast collision detection inference of around 7$\mu$s/query and is 19.7$\%$ mAP higher than baselines when tested on 2.5 million queries in five diverse sets of environments. Using the same \cabinet encoding, we learn a scene SDF-based waypoint sampler and show that it is crucial for transitioning between pick and place actions.
    \item We demonstrate zero-shot \textit{sim2real} transfer for our model on completely unknown scenes and objects in the real world, including out-of-distribution kitchen environments.
\end{itemize}

%% file: 02_related_work.tex
\section{Related Work}

\textbf{Collision Detection from Point Clouds:} There are a variety of options to check for collision with known object meshes or fully visible point clouds. One can use computational geometry libraries~\cite{fcl2012} if the object mesh is known. Alternatively, one can voxelize or spherize~\cite{hubbard1996approximating} the point clouds and formulate the collision checking problem as evaluating whether any of the elements is in collision with robot links. However, voxel-based approaches suffer from occlusion which is mitigated through use of multiple views. This unfortunately constraints the robot workspace or requires mapping of the the environments which needs to be dynamically updated as objects are moving around. SceneCollisionNet~\cite{Danielczuk2021ObjectRU} frames the collision checking problem as a hybrid of classical voxel based method and data driven methods. It encodes the scene to coarse voxels where each voxel is represented by a deep embeddding vector. Each collision query is defined as a pair of query object and scene point clouds. Collision checking is done with a binary classifier which takes as input the scene voxel embedding, object embedding, and the relative SE$(3)$ transformation to that voxel. SceneCollisionNet was trained on only the table top scenes. In this paper, we build on top of SceneCollisionNet. By scaling up the training data to go beyond table top settings, we observe a boost in performance and generalization across variety of different scenes.

\textbf{Neural Rearrangement Planning:} Traditionally, solutions to object rearrangement have been dominated by model-based methods, such as TAMP~\cite{garrett2021} which use an explicit 3D world representation estimated from sensor observations.
More recently, there has been an increase in learning-based vision-centric rearrangement approaches~\cite{batra2020rearrangement}, although the majority assumes simplified action spaces that abstract away the actual grasping and placing motion and often focus on navigation.
In~\cite{labbe2020monte} a learned visual state estimator is combined with a Monte-Carlo tree search planner, to efficiently solve planar tabletop rearrangements.
When goals are provided as target images, transporter networks~\cite{zeng2020transporter}, their equivariant version~\cite{huang2022equivariant} or goal-conditioned transporter networks~\cite{wu2022transporters} can be used for pick-and-place.
The problem of matching object instances between goal and initial image can also be simplified with vision-language models~\cite{goodwin2022semantically}.
Closest to our approach for rearrangement are NeRP~\cite{nerp2021} and IFOR~\cite{ifor2022}.

\textbf{Large-scale Procedural Scene Generation:} Probabilistic models for indoor scene generation have been originally developed in computer graphics~\cite{fisher2012example,majerowicz2013filling,yu2015clutterpalette}.
Apart from appealing visually, simulating scenes physically requires more care when arranging objects.
As a result most data available for learning robot manipulation in simulation is limited to a fixed number of artist designed scenes:~iGibson~\cite{li2021igibson}, Meta-World~\cite{yu2019meta}, RLBench~\cite{james2019rlbench}, Sapien~\cite{Xiang_2020_SAPIEN}, Habitat~\cite{szot2021habitat}, or AI2 ManipulaTHOR~\cite{ehsani2021manipulathor}.
Those scenes mostly consist of assets and arrangements from datasets such as PartNet~\cite{Mo_2019_CVPR}, ReplicaCAD~\cite{replica19arxiv}, and 3D-Front~\cite{fu20213d}.
More recently, ProcTHOR~\cite{deitke2022procthor} has shown to procedurally generate large amounts of entire apartment layouts with room-specific object arrangements. Nevertheless, our use case focuses on smaller-scale clutter for which ProcTHOR’s scenes are not dense enough. We will present our procedural scene generation pipeline next.

%% file: 03_method.tex
\section{Procedural Data Generation}
\label{proc_data}

\textbf{Synthetic Scene Generation:} We procedurally generate synthetic data in simulation. To generate our cluttered scenes, we first assemble a set of environment assets $\mathcal{E}$, object assets $\mathcal{O}$ and fixed robot manipulator $\mathcal{R}$ (Franka Panda in our case). We have a probabilistic grammar \cite{kar2019metasim} $P$ which dictates how the assets can be organized into random scene graphs $\mathcal{S}\sim P(\mathcal{E},\mathcal{O},\mathcal{R}) $. This grammar is composed of the following key components: 1) sampling potential supports surfaces $\gamma$ in an environment asset to place objects 2) rejection sampling to sequentially place objects on these surfaces without colliding with the scene and 3) fixing the robot base in a region where there is sufficient intersection over union (IoU $>$ 0.8) between the workspace of the robot (approximated by a cuboid volume) and $\gamma$. Once the scene $S$ is generated, collision queries are sampled with free-floating object meshes (computed in a straight-line trajectory) and the scene. 
The synthetic point clouds $X$ are rendered online during training.

\begin{figure}
  \begin{center}
    \includegraphics[width = 0.95\linewidth]{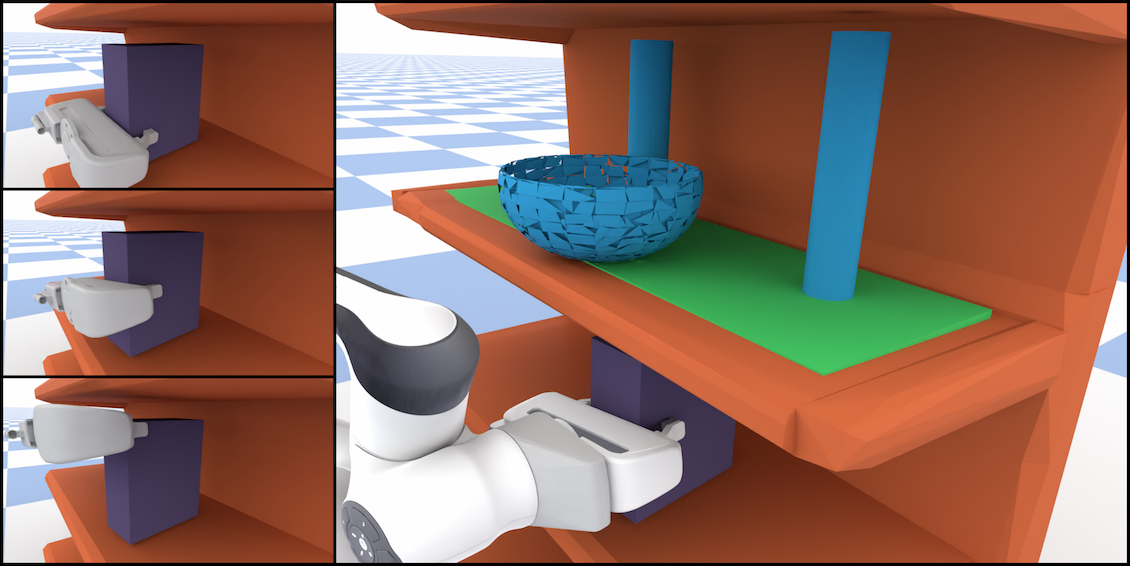}
  \end{center}
  \caption{An example of a procedurally generated CabiNet rearrangement scene. The target object (here in purple) is chosen if it has a selection of valid collision-free grasp poses. The green region represents the placement shelf, which is chosen if a) it is different from the shelf the object originates from and b) has a valid placement pose for the target object.}
  \vspace{-5mm}
\label{fig:rearrangement_prob}
\end{figure}

\textbf{Dataset:} Our dataset of object assets for training comes from ACRONYM \cite{acronym2020}, that contains wide range of object geometries from 262 categories as well as high-quality $SE(3)$ grasps which we use for picking objects. We split the dataset for training and testing. Fig~\ref{fig:cover} depicts examples of our environment assets, which we chose from common categories such as shelves, cubby, cabinet, drawers and table. All the assets are procedurally generated with the exception of shelves. For shelves, we aggregate the shelf categories from ShapeNetCore\cite{shapenet} and filter assets which cannot be made watertight or if proper support surfaces cannot be extracted. Nonetheless, the object placements on all the environments, including the shelves, are procedurally generated. We include more examples of scenes in Appendix \ref{appendix:more-scenes}. For testing, we only consider assets from the shelf environment dataset and the objects are from a novel dataset \cite{6dofgraspnet} unseen during training. More exa

\textbf{Rearrangement Problem Generation:} A valid rearrangment problem needs a scene, a target object that the robot needs to grasp and the placement shelf that the robot needs to place the object. Given a scene, a target object is sampled if there exists a set of ground truth grasps associated with that object where they do not collide with the environment and have valid collision free inverse kinematic configuration. Once the target object is sampled, we check if a collision free placement pose for the target object exists within the placement shelf. To make sure that our rearrangement problems are challenging, the placement shelf is going to be different from the shelf that has the pick object. A problem is chosen if it passes both stages of sampling target object and also having a valid placement location. Overall, this process has a success rate of $20.8\%$ in finding successful problems. An example of rearrangement problem is shown in Fig~\ref{fig:rearrangement_prob}.

\section{Neural Rearrangement Planning}
\begin{figure*}[t]
\centering
    \includegraphics[width=0.95\linewidth]{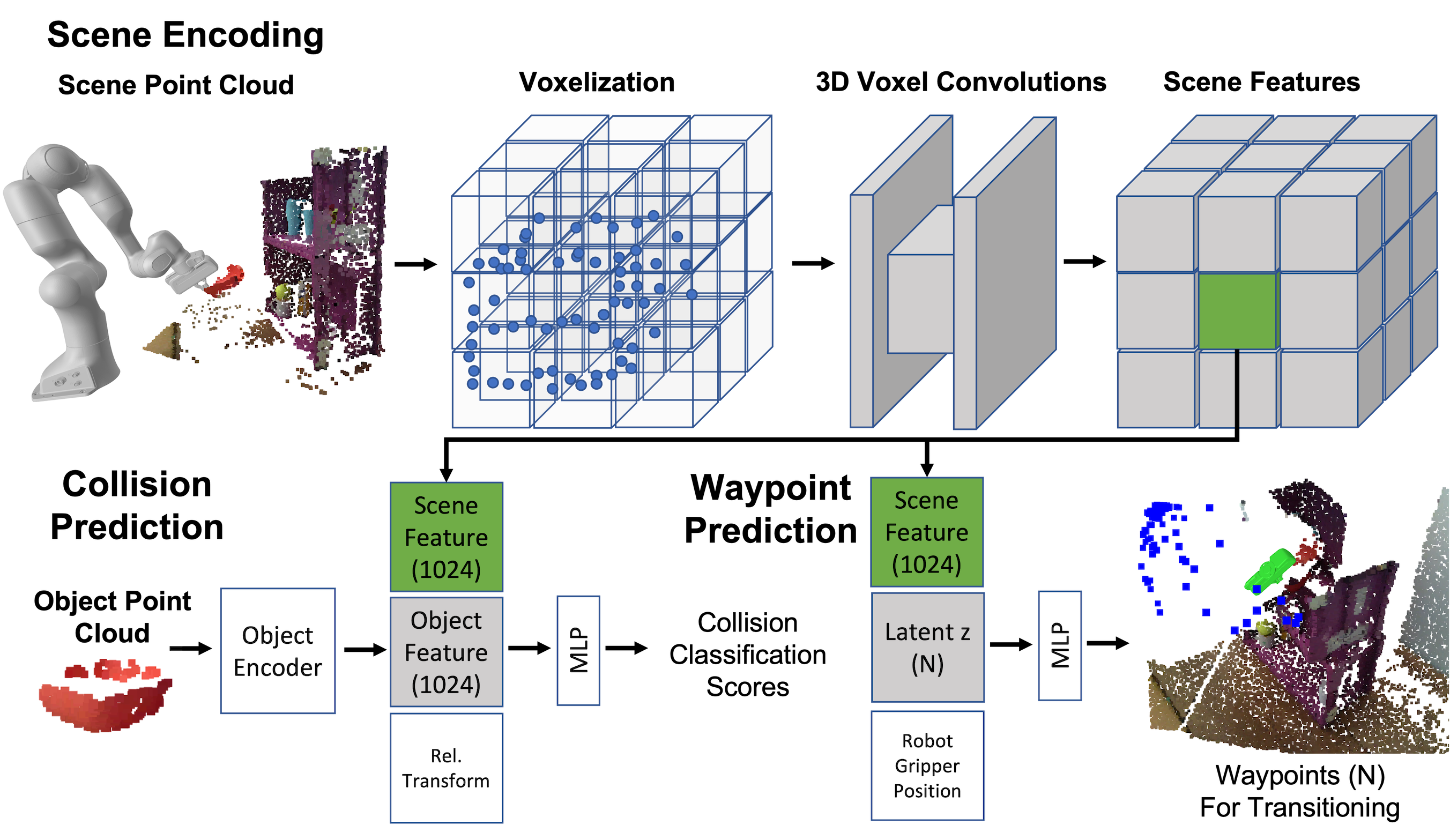}
  \caption{Our \cabinet architecture first encodes the scene point cloud with voxelization and 3D convolutions, shown in the top. The robot is only used for visualization and the robot point cloud is removed from the scene in practice. The scene features are then used with the object features to predict scene-object collision queries. We also predict waypoints (points colored in blue) for rearrangement, conditioned on latent vector $z$ and the current gripper position (shown in green).}
  \vspace{-5mm}
\label{fig:architecture}
\end{figure*}

Our approach and model architecture is summarized in Fig \ref{fig:architecture}. We first learn a implicit 3D encoding of the scene point cloud. We use the encoded scene feature along with learned object features for fast point-cloud based collision detection with \cabinet. The same scene feature is then used for predicting waypoints for the rearrangement task with a simple feedforward network. Both these models are then used to generate robot trajectories for object rearrangement with a Model Predictive Path Integral (MPPI) policy \cite{mppi2017}. More details of the trajectory generation process is given in Appendix \ref{appendix:traj_gen}.

\textbf{Collision Prediction:} This is a learned collision model that accepts as input the scene point cloud $X_{S}$ and the object point cloud $X_{O}$. The point cloud is encoded with voxelization and 3D convolution layers $\Psi_{S} = Enc(X_{S})$. This is followed by a MLP binary classifier $c = g_{\theta}(\Psi_{S}, \Psi_{O}, T_{O \rightarrow S})$ that predicts if the object collide with the scene, where $T_{O \rightarrow S}$ is the relative transformation between the object and the scene and $\Psi_{O}$ are the object features encoded with PointNet++ \cite{pointnet} layers akin to \cite{Danielczuk2021ObjectRU}. We adopt the model architecture of prior work \cite{Danielczuk2021ObjectRU} but it was only trained on table top scenes which results in poor generalization to other type of scenes such as shelves. We showed that by scaling up the training data to more diverse set of environments the model generalizes to different type of scenes. \cabinet is trained with binary cross entropy loss and with SGD with constant learning rate. Overall, we train \cabinet for two weeks, considering over 650K scenes and for 60 billion query pairs. We also modified the architecture of \cite{Danielczuk2021ObjectRU} by increasing the voxel sizes.

\textbf{Waypoint Prediction:} Object rearrangement in more constrained environments such as shelves imposes new challenges. Some of the approaches that work quite reliably in table top settings fail in navigating the tight spaces between shelves. One such approach is the work of \cite{Danielczuk2021ObjectRU} that finds the collision free path by rolling out multiple trajectories in configuration space ($C$-space) between the current configuration of the robot and the closest goal $\mathcal{G}$ in the $C$-space. These rollouts are sampled around a straight line that connects the current robot configuration to $\mathcal{G}$. Given the nominal line between the robot configuration and $\mathcal{G}$, different lines are sampled where the slope of the lines are gaussian distribution centered at the slope of nominal trajectory with a predefined variance. Each rollout is trimmed at the point of collision using SceneCollisionNet and ranked based on the distance of the rollout's final point to $\mathcal{G}$. The success of this approach hinges on the quality of the sampled trajectory. The simple sampling explained above, works quite well for table top scenes where there is more free space. However, it fails to sample promising trajectories for shelves and more constrained environments, such as when the robot needs to move from one compartment of the shelf to another. The majority of the sampled trajectories around the nominal trajectory would go through the divider between shelves. 

To address this shortcoming, we propose to use \cabinet to sample waypoints with a larger signed distance value. Given the current end-effector position of the robot and the scene point cloud, it samples goals that gets the robot out of tight spaces. More formally, these waypoints $w \in \mathbb{R}^{3}$ are defined as to be in the set $\{ w| \tau_{min} \leq SDF(S, w) \leq \tau_{max} \cap \lVert w - p_{gripper} \rVert \leq D\}$, where $S$ is the scene mesh, $p_{gripper}$ is the end-effector position. The \cabinet waypoint sampler is modelled as a conditional generator $\hat{w} = f_{\theta}(\Psi_{S}, p_{gripper}, z)$ as in Generative Adversarial Networks \cite{gan2014}. 
Instead of using adversarial training, we train the sampler with Implicit Maximum Likelihood Estimation (IMLE) \cite{imle2018} which attempts to make each generated sample similar to a ground truth sample. We empirically found that making the loss bidirectional improved the generated samples - enforcing that ground truth samples are also similar to the set of nearest predicted samples. Let $z_{1}, \ldots, z_{m} \sim \mathcal{N}(0, I)$ denote randomly sampled latent input noise vectors and $w_{i}$ the ground truth waypoints. The IMLE loss is as follows:

\vspace{-3mm}

\begin{equation}
\mathcal{L}_{IMLE} = \frac{1}{n} \sum_{i=1}^{n} \min_{j} |\hat{w_{j}} - w_{i}| + \frac{1}{m} \sum_{j=1}^{m} \min_{i} |\hat{w_{j}} - w_{i}| 
\label{eq:imle}
\end{equation}

In our experiment, we let $\tau_{min}=0.40$, $\tau_{max}=0.45$ , $D=0.40$ and both $m$ and $n$ are 70. We let $z$ have two dimensions and train with SGD with a constant learning rate. At both training and inference, the latent vectors are sampled from $z \sim \mathcal{N}(0, I)$.

\textbf{Object Rearrangement: } The rearrangement process can typically be broken into a sequence of the following three states: \textit{pick}, \textit{transition} and \textit{place}. For the picking action, we use Contact-Graspnet~\cite{sundermeyer2021contact} to sample a batch of 6-DOF grasps for the target object. For placing objects, we first sample potential object positions based on the scene point cloud and the support surface. The placement orientation is set to the current object pose while it is being grasped. The poses are filtered by \cabinet for collisions with the environment and followed by whether a kinematic solution can be found for the manipulator. These 6-DOF poses are then used to construct motion planning problems used to generate robot trajectories with MPPI, as described in Section \ref{appendix:traj_gen}.

%% file: 04_results.tex
\section{Experimental Evaluation}

\subsection{Evaluation on Collision Benchmark}
We evaluate \cabinet on a collision benchmark against four baseline point cloud-based collision detection algorithms. We want to emphasize that our setting only requires a point cloud observation from a single view. We sample synthetic scene/object point cloud pair where the objects move in 64 linear trajectories in a scene. The collision ground truth information is computed with FCL \cite{fcl2012} in simulation. For each experiment, we have a balanced set of 256K collision and collision-free queries for a total of 512K queries/experiment. We evaluated on five environments (1000 scenes each) from Fig \ref{fig:cover} and the results are averaged across them in Table \ref{tb:collisionnet}.

\textbf{Quantitative Metrics:} We report the following 1) mean Average Precision (mAP) score for the classifier, averaged across the five environment test sets 2) collision prediction accuracy and 3) time/query in $\mu$s. Our baselines are as follows:
\begin{itemize}
    \item \textbf{SceneCollisionNet} \cite{Danielczuk2021ObjectRU}: We directly evaluated the pretrained model from prior work. This approach was just trained on a single environment (Tabletop) with a fixed robot-to-tabletop transformation, and directly infers collision from point clouds without any preprocessing.
    \item \textbf{Occupancy Mapping} \cite{chitta2012moveit}: This is one of the most commonly used geometric collision checking heuristic representation used in the robotics community \cite{chitta2012moveit, mbm2021, octomap}. We use the open source implementation from Open3D \cite{Zhou2018} to convert the sensed point cloud to a occupancy map representation. Voxels are specified to be of 1$cm$ in size and are labelled to be either collision free or occupied.
    \item \textbf{Marching Cubes + FCL} \cite{fcl2012}: We first convert the scene and object point clouds to a mesh with the marching cubes algorithm and use FCL to compute the collision between the scene and object meshes. We parallelize this baseline across 10 processes for a fairer comparison.
    \item \textbf{Marching Cubes + SDF} \cite{KaolinLibrary}: The scene point cloud is first converted to a mesh and we fit a SDF to it using the GPU implementation from \cite{KaolinLibrary}. Each point in the object point cloud is computed for its SDF value from the scene. If any of the points have a negative distance (due to a penetration), the entire scene/object point cloud is considered to be in collision.
\end{itemize}

\textbf{Baseline Comparisons:} Overall \cabinet outperforms the baselines methods in terms of both accuracy and inference speed. It has the highest mAP across the five environment test sets. It is nearly 24$\%$ and 15$\%$ higher mAP and accuracy respectively compared to OccupancyMap \cite{chitta2012moveit} which is a popular method in the community, while being nearly 25x faster with a 7$\mu$s inference time. OccupancyMap performance has direct correlation with the coverage of point cloud over occluded part of the scenes. The more occluded areas in the scene, the less accurate it becomes. \cabinet, on the other hand, does not suffer from the occlusion issue since it is trained with single camera and has been learned to extrapolate to occluded parts in order to solve collision queries.  \cabinet also generalizes to more diverse environments and point data compared to the pretrained SceneCollisionNet \cite{Danielczuk2021ObjectRU} which shows the importance of training on diverse set of scenes and objects. Our approach is also about 4X faster than the parallelized FCL baseline with a 20.4$\%$ higher mAP score.

\textbf{Ablation on Environments:} We show in Table \ref{tb:generalization} that \cabinet generalizes to diverse environments by training with more in-distribution data. We notice that the model generalizes to similar environments even without any training, such as cabinets and shelves. The model trained on all the environments performed the best on all the test sets.

\begin{table}
\footnotesize
\centering
\caption{Results on Collision Benchmark}
\label{tb:collisionnet}
\begin{tabular}{c|ccc}
\hline
 \textbf{Collision Model} & \textbf{mAP}  & \textbf{Accuracy} (\%) & \textbf{Time/Query ($\mu$s)}    \\
\hline
\cabinet (Ours) & \textbf{0.971} &  \textbf{89.0} & \textbf{6.41 $\pm$ 3.58}   \\
SceneCollisionNet \cite{Danielczuk2021ObjectRU} & 0.706 & 69.9 & 7.03 $\pm$ 3.89  \\
OccupancyMap \cite{chitta2012moveit} & 0.732 & 74.1 & 174.5 $\pm$ 61.9  \\
MC + FCL \cite{fcl2012} & 0.767 & 78.8 & 27.5 $\pm$ 6.82 \\
MC + SDF \cite{KaolinLibrary} & 0.773 & 80.0 & 168.6 $\pm$ 46.4 \\
\hline
\end{tabular}
\end{table}

\begin{table}
\footnotesize
\centering
\caption{\cabinet Generalization to Environments}
\label{tb:generalization}
\begin{tabular}{c|cccccc}
\hline
  \multirow{2}{*}{\textbf{Train Set}} & \multicolumn{6}{c}{\textbf{Test Set (AP)}} \\
  \cmidrule(lr){2-6} & Tabletop & Shelf & Cubby & Drawers & Cabinet & \textbf{mAP} \\
\hline
Tabletop & \textbf{0.989} & 0.910 & 0.855 & 0.861 & 0.855 & 0.894 \\
Shelf & 0.924 & \textbf{0.985} & \textbf{0.978} & 0.956 & 0.972 & 0.963 \\
Cubby & 0.930 & 0.974 & 0.977 & 0.961 & 0.990 & 0.966 \\
Drawers & 0.923 & 0.856 & 0.848 & \textbf{0.972} & 0.914 & 0.903 \\
Cabinet & 0.924 & 0.971 & 0.961 & 0.961 & \textbf{0.990} & 0.961 \\
\hline
All Envs & 0.971 & 0.969 & 0.965 & 0.971 & 0.979 & \textbf{0.971} \\
\hline
\end{tabular}
\end{table}

\begin{table*}
\footnotesize
\centering
\caption{Simulated Rearrangement Experiments}
\label{tb:waypoint}
\begin{tabular}{cc|ccc}
\hline
\multirow{2}{*}{\textbf{Planner}} & \multirow{2}{*}{\textbf{Collision Model}} & \multicolumn{3}{c}{\textbf{Waypoint Type}}    \\
\cmidrule(lr){3-5} & & \cabinet (Ours) & Reverse Approach & No Waypoints \\
\hline
MPPI & \cabinet (Ours) & \textbf{36.6\%/159s} & 16.7\%/158s & 4.3\%/201s \\
MPPI & OccupancyMap \cite{chitta2012moveit}  & 15.0\%/289s & 11.0\%/234s & 7.5\%/307s \\
AIT* \cite{Gammell2015BatchIT, Hauser2010FastSO} & OccupancyMap \cite{chitta2012moveit} & 21.0\%/808s & 18.5\%/380s  & 5.0\%/451s  \\
RRTConnect \cite{rrtconnect} & OccupancyMap \cite{chitta2012moveit} & 26.4\%/357s & 18.1\%/380s  & 5.8\%/389s  \\
\hline
\end{tabular}
\vspace{-4mm}
\end{table*}

\begin{table}
\footnotesize
\centering
\caption{Success Rate by States}
\label{tb:state-machine}
\begin{tabular}{c|cccc}
\hline
\multirow{2}{*}{\textbf{Waypoint Strategy}} & \multicolumn{4}{c}{\textbf{States}}    \\
\cmidrule(lr){2-5} & Pick & Transition & Place & Overall \\
\hline
\cabinet (Ours) & \textbf{61.0\%} &	\textbf{80.0\%}	& \textbf{75.0\%} & \textbf{36.6\%} \\
Reverse Approach &  54.8\% &	43.5\%	& 70.0\% & 16.7\% \\
No Waypoints & 60.9\% & 21.4\% & 33.3\% & 4.3\% \\
\hline
\end{tabular}
\end{table}

\subsection{Object Rearrangement Evaluation in Simulation}
We evaluate our collision model in simulated rearrangement trials in IssacGym \cite{issacgym2021} as shown in Table \ref{tb:waypoint}. To focus more on the rearrangement aspect of the problem, we used the ground truth grasp poses from \cite{6dofgraspnet}. We adopt a standard state-machine for rearrangement tasks used in prior works \cite{danielczuk2019, nerp2021, ifor2022}.
The objects are chosen from bowl, box, and cylinder categories of \cite{6dofgraspnet} and are held out from training data.
For the environment assets, we use the seven shelves (from ShapeNet) in our test set to construct 30 scenes. We only use one fixed scene camera for all experiments and each scene gets two rearrangement trials, for a total of 60 experiments for each method. 

\textbf{Quantitative Metrics:} We report three metrics on this task: 1) \emph{overall success rate} is the success rate for the whole pick and place operation where the robot picks the target object and place it in the designated shelf without any failures. 2) \emph{individual success rate} is the success rate of each state given the number of times the policy state machine reaches to each particular state and 3) \emph{total time} taken for each experiment. There are three stages in rearrangement: pick, transitioning to a placing pose and place. A pick is considered a success if the object is grasped after lifting the object from the support surface. The same is true for transition. The placement is considered a success if the object is detected to be resting on the place support surface, regardless of its final orientation. We emphasize that rearrangement is a long-horizon task and conditional success rates for each state are based on the performance of the previous state. As a result, even if the performance of individual state success rates are high, errors accumulate over time leading to a lower overall success rate.

\textbf{Collision Representation for Planning:} We compared to Occupancy Mapping since it a popular heuristic collision representation used in the community\cite{chitta2012moveit}. The performance of all planners using this collision model significantly deteriorated compared to our \cabinet model. This shows the benefit of data driven approaches where they reason beyond the part of the point cloud that is visible and implicitly reason about occlusions as well. Occupancy maps is also significantly slower, increasing the rearrangement time by about 80\% for the MPPI planner using the \cabinet waypoint sampler.

\textbf{Comparison to Global Planning pipeline:} We compare to the standard off-the-shelf motion planning pipeline used in the community with a Occupancy Mapping \cite{chitta2012moveit} representation. Specifically, we compare to the state-of-the-art configuration space planner \cite{Gammell2015BatchIT}, which is an almost-surely asymptotically optimal planner. We also compare to RRTConnect\cite{rrtconnect}, which commonly is used to find feasible, though not necessarily optimal, paths. We give a timeout of 60$s$ to find a solution for both planners. After planning, we apply spline-based, collision-aware trajectory smoothing~\cite{Hauser2010FastSO} to the solutions. If the planner fails to find a valid solution, we simply execute the greedy solution to the goal to continue with the rearrangement process. Overall, the MPPI planner with our \cabinet model outperforms RRTConnect and AIT* by about 10$\%$ and 15$\%$ respectively and is also significantly faster.

\begin{figure}
  \begin{center}
    \includegraphics[width = 0.95\linewidth]{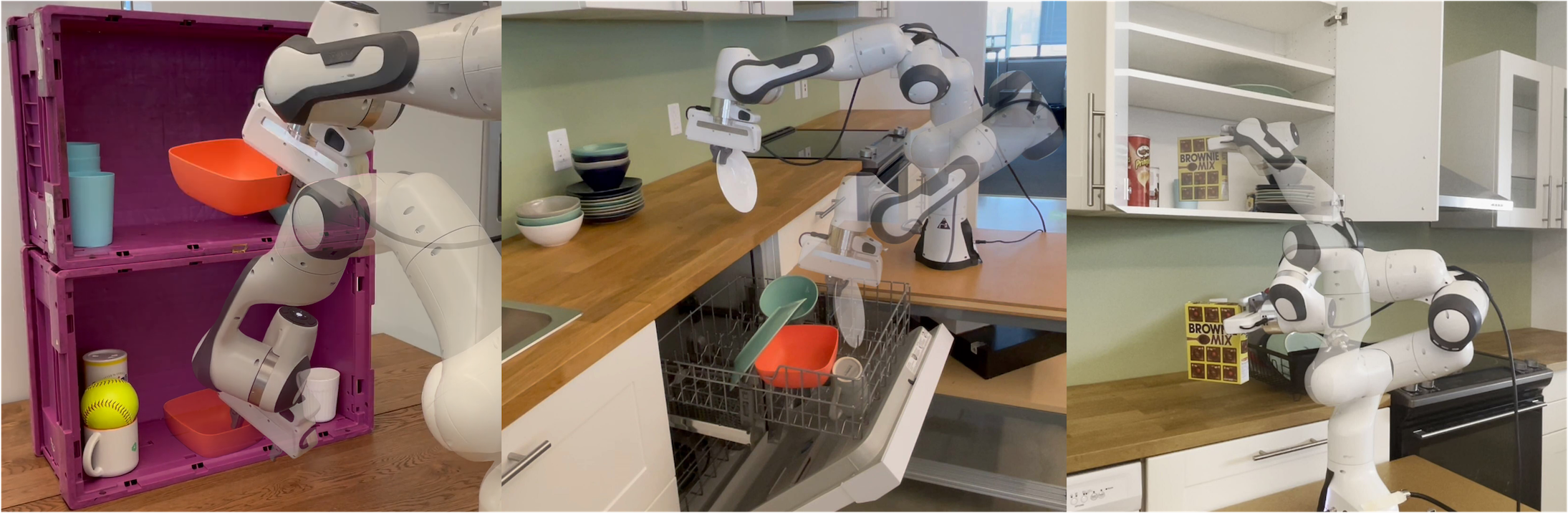}
  \end{center}
  \caption{\cabinet is used for pick-and-place tasks in the real world. The scenes in the middle and the right are out-of-distribution environments in a real IKEA kitchen.}
\label{fig:real_robot}
\end{figure}

\textbf{Importance of Waypoints:} We demonstrate that learning waypoints are crucial for rearrangement to navigate out of tight spaces. We compare to a common heuristic used in the motion generation literature to move robots out of tight spaces \cite{9682604}, which is to move the gripper in the reverse of the approach direction. We noticed that this approach, while proficient for primitive shapes (e.g. cylinders) it does not scale to more complex shapes like bowls, which have more complicated 6-DOF grasps. Hence, retracting in the reverse of the approach direction with an object in hand, the grasped object could collide with neighbouring support surfaces while transitioning from the pick to the place poses. As shown in Table \ref{tb:state-machine}, our \cabinet waypoint sampler improves the transition success rate by nearly 60$\%$ compared to when having no waypoints and about 35$\%$ when compared to the strategy of retracting in the opposite of the approach direction.

\subsection{Real Robot Experiments}

We run experiments to show that our model transfers to a real robot despite only being trained in simulation.

\textbf{Hardware Setup:} Experiments are done on a 7-DOF Franka Panda Robot with a parallel-jaw gripper. The system is equipped with two cameras: 1) Wrist mounted camera, which is an Intel Realsense D415 RGB-D camera, that is used for grasping 2) External camera, which is an Intel L515 RGB-D camera, that is used for generating the point cloud for \cabinet. Grasps are generated using Contact-GraspNet~\cite{sundermeyer2021contact} and placement shelf is manually annotated for each problem by labeling the region that belongs to the desired placement shelf.
We use the model from \cite{objseeker} to run instance segmentation on both cameras. The user selects the target object by clicking on the external camera image. Upon user selection of target object, the robot takes a closer look at the object and find the object in the wrist camera through relative camera pose between wrist camera and external camera. This step is crucial for grasp success since the wrist camera provides denser points on the object and it mitigates the effect of calibration imperfections.  \cabinet has access to only the external camera point cloud and the inference is run on NVIDA Titan RTX gpu.

\textbf{Experiment Setup:} We test our approach in novel environments, with unseen shelf and objects assets in unknown poses.  We experiment with three objects from different object categories and two tasks. As shown in Fig \ref{fig:real_robot} in the vertical transport task, the robot has to pick an object from the top shelf and place it in the bottom one, or vice versa. Similarly for horizontal transport task the shelf compartments are horizontally next to each other. For each task-object pair, we attempt four trials, two of which go from one support compartment to the next and two in the opposite direction. In total we have 24 experiments and each task-object pair has a unique environment, where  the object has to be picked and placed amidst clutter.

\begin{figure}
  \begin{center}
    \includegraphics[width = 0.89\linewidth]{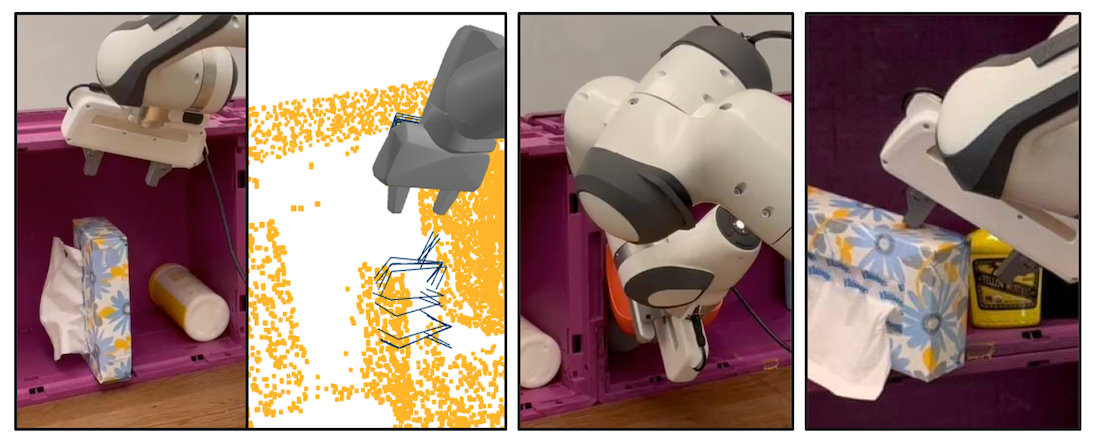}
  \end{center}
  \caption{Examples of failure cases, left: roof partially occluded leading to collision with wrist camera during grasping, middle: grasped object collided with barrier, right: pick failure.}
\label{fig:failure}
\end{figure}

\textbf{Discussion:} Results are reported in Table \ref{tb:robot} and the vertical and horizontal transfer tasks have 75$\%$ and 58$\%$ success rates respectively. There were 2/24 pick failures due to incorrect grasps. One attempt failed during placing and 5/24 attempts failed when transitioning from pick to place states. The horizontal transfer task was relatively more challenging due to two reasons - the transitioning was over a longer distance leading to a greater chance of collision and the leftmost shelf compartment was not entirely visible from our static scene camera. Three failures were specifically due to occlusion and the \cabinet model not seeing enough of the leftmost cubby geometry from our camera setup, as shown in Fig \ref{fig:failure}. The real robot executions are included in the \href{https://youtu.be/Bs5QYkJVcjM}{supplementary video}.

\textbf{Out-Of-Distribution Environments:} We were able to deploy the \cabinet model for pick-and-place tasks in out-of-distribution environments in a real IKEA kitchen (shown in Fig \ref{fig:real_robot}), such as grabbing a plate from a dishwasher and grasping a box from a cabinet. The scene camera was extrinsically re-calibrated for each environment. We hypothesize that a combination of large-scale training in simulation and inductive biases in our architecture (including the use of camera calibration and point cloud representation) allowed \cabinet to easily generalize to these disparate environments.

\begin{table}
\footnotesize
\centering
\caption{Real Robot Experiments}
\label{tb:robot}
\begin{tabular}{c|cccc}
\hline
\multirow{2}{*}{ \textbf{Task}} & \multicolumn{3}{c}{\textbf{Object Category}}    \\
\cmidrule(lr){2-4} & Bowl & Box & Bottle & Overall \\
\hline
Vertical Transport & 75.0\% & 50.0\% & 100.0\%  & 75.0\% \\
Horizontal Transport & 50.0\% & 25.0\% & 100.0\%  & 58.0\% \\
\hline
\end{tabular}
\vspace{-2mm}
\end{table}

%% file: 05_conclusion.tex
\section{Conclusion}
We present an effort in scaling up neural rearrangement in clutter. We train our \cabinet model to predict collisions and motion waypoints from point cloud observations. It outperforms baseline approaches in terms of collision predicted, simulated experiments and also transfers well to real world clutter despite being only trained in simulation. While we perform our simulated and real experiments on a franka robot, we want to emphasize that the \cabinet model is conceptually robot-agnostic and can potentially work with other robots without re-training. A limitation of our architecture is that the 3D voxelization enforces queries to be within the model workspace which can sometimes be out of bounds during manipulation. Potential extensions could also explore more complicated scenes or learning to generate synthetic scenes \cite{kar2019metasim}. One could also explore hybrid architectures leveraging recent learned motion policies \cite{fishman2022mpinets} that are faster than traditional planners, along with \cabinet. Videos of robot experiments are available at: \href{https://cabinet-object-rearrangement.github.io}{https://cabinet-object-rearrangement.github.io} and the supplementary material are in the appendix.

%% file: 06_acknowledgements.tex
\section*{Acknowledgements}
We would like to thank Ankur Handa and Jan Czarnowski for discussions and inspiring the project name; Wei Yang and Yu-Wei Chao for helping with Omniverse rendering and camera calibration; Balakumar Sundaralingam, Lucas Manuelli, Chris Paxton and ‪Towaki Takikawa for discussions on the project.

%% file: appendix.tex
\appendix
\section*{Appendix}

We provide more examples of the synthetic data we used for training in Section \ref{appendix:more-scenes}. We also describe how the \cabinet model was used in generating robot trajectories for object rearrangement in \ref{appendix:traj_gen}.

\section{Additional Examples of Synthetic Data and Scenes}
\label{appendix:more-scenes}

\begin{figure*}[!ht]
  \centering
  \includegraphics[width=\textwidth]{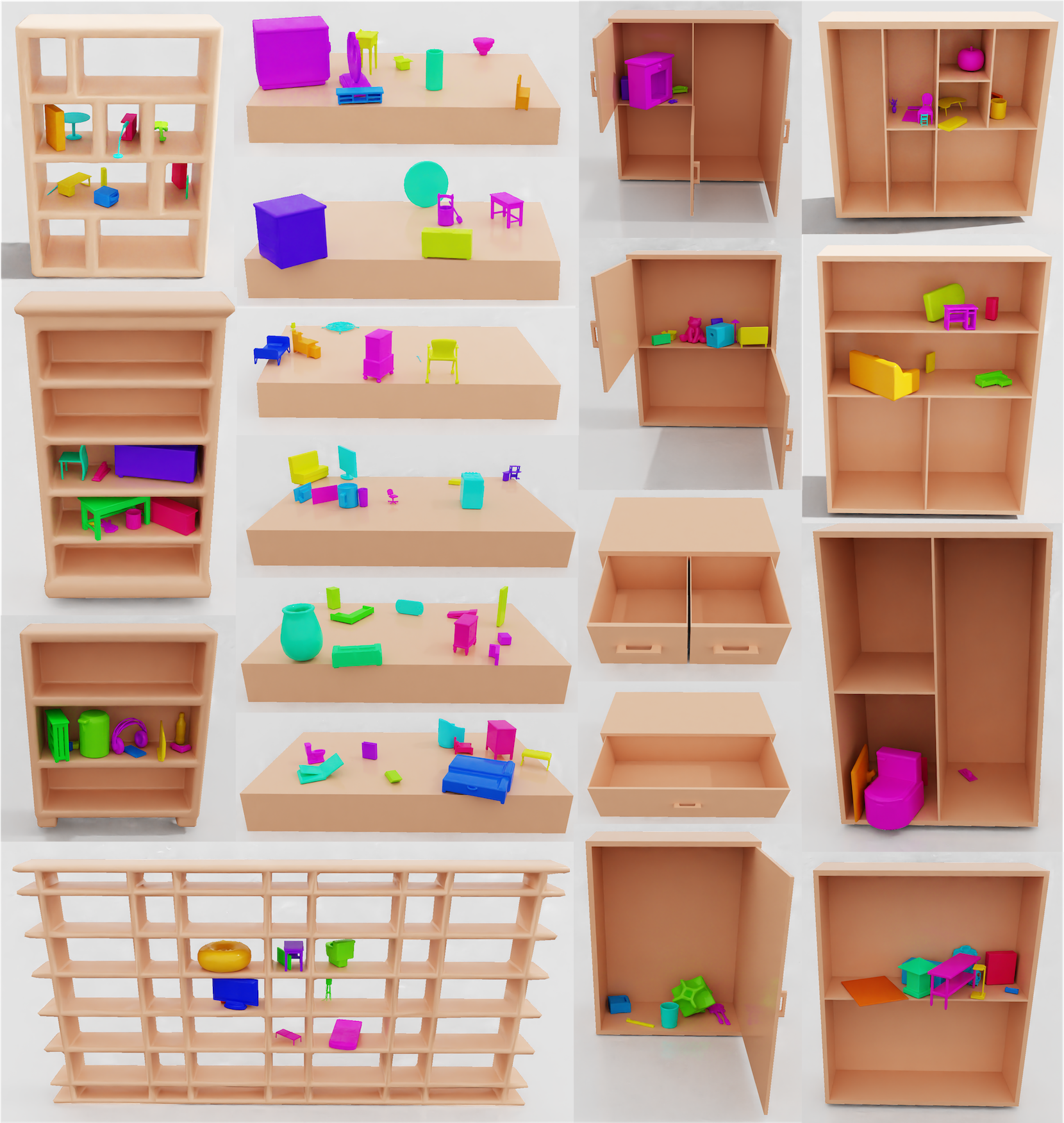}
  \caption{We procedurally sampled scenes from five distribution of environments (from left to right): shelf (from ShapeNet), tabletop,  cabinet, drawer and cubby. The scenes are exported to the USD file format and rendered in  \href{https://www.nvidia.com/en-us/omniverse/}{NVIDIA Omniverse}. See Section \ref{proc_data} for more details.}
  \label{fig:more_dataset}
\end{figure*}

\section{Trajectory Generation}
\label{appendix:traj_gen}

We refer the authors to the SceneCollisionNet \cite{Danielczuk2021ObjectRU} paper for a more detailed treatment of the trajectory generation process. We use a Model Predictive Control (MPC) framework to generate trajectories for object rearrangement. We leverage the GPU-parallelism inherent in the \cabinet architecture with batch collision queries. The task is specified entirely in configuration (joint) space and constraints in the form of joint limits, self-collisions and robot-scene collisions are enforced. With the exception of inverse kinematics (for which we use a CPU implementation with multi-processing), the entire trajectory generation stack consisting of trajectory sampling, cost computation, forward kinematics and collision checking is tensorized to run on a GPU for fast inference and closed-loop execution. We use a simple heuristic for sampling trajectories which we then filter using rejecting sampling. We construct a straight line trajectory $d$ connecting the start and goal joint configuration and sample $\tau$ trajectories drawn from a normal distribution centered on $d$: $\hat{d}=N(\mathcal{N}(d, \Sigma))$. We renormalize the direction vector $\hat{d}$ and construct trajectories $\tau$ with $H$ steps along $\hat{d}$. The cost of each trajectory is based on distance in configuration space and lowest cost trajectory is then executed on the robot with a position controller.

The sampled robot trajectories are filtered with rejecting sampling. We first enforce joint limit constraints by clipping the trajectories and check for self-collisions at each step of the trajectory using the model trained in \cite{Danielczuk2021ObjectRU}. For the robot-scene collisions, we use the \cabinet model. We sample point clouds from the surface of the collision mesh, for each link of the robot manipulator and treat it as a object point cloud when constructing the collision queries. For each trajectory, we make $H \times D$ collision queries as a single forward pass in \cabinet, where $D=7$ is the number of links for the franka robot. We empirically found that the \cabinet model could be used to sample robot-scene collision queries for the franka robot model even though it was not included in the training dataset. This demonstrates the generalization capability of the architecture since it trained with large scale synthetic data.

%% file: 00_main.bbl
\begin{thebibliography}{10}

\bibitem{octomap}
M.~Bennewitz C.~Stachniss A.~Hornung, K. M.~Wurm and W.~Burgard.
\newblock Octomap: An efficient probabilistic 3d mapping framework based on
  octrees.
\newblock In {\em Autonomous Robots}, 2013.

\bibitem{batra2020rearrangement}
Dhruv Batra, Angel~X Chang, Sonia Chernova, Andrew~J Davison, Jia Deng, Vladlen
  Koltun, Sergey Levine, Jitendra Malik, Igor Mordatch, Roozbeh Mottaghi,
  Manolis Savva, and Hao Su.
\newblock Rearrangement: A challenge for embodied ai.
\newblock 2020.

\bibitem{robotiq2021}
Mathieu Belanger-Barrette.
\newblock What is an average price for a collaborative robot?
\newblock 2021.

\bibitem{rt12022arxiv}
Anthony Brohan, Noah Brown, Justice Carbajal, Yevgen Chebotar, Joseph Dabis,
  Chelsea Finn, Keerthana Gopalakrishnan, Karol Hausman, Alex Herzog, Jasmine
  Hsu, Julian Ibarz, Brian Ichter, Alex Irpan, Tomas Jackson, Sally Jesmonth,
  Nikhil Joshi, Ryan Julian, Dmitry Kalashnikov, Yuheng Kuang, Isabel Leal,
  Kuang-Huei Lee, Sergey Levine, Yao Lu, Utsav Malla, Deeksha Manjunath, Igor
  Mordatch, Ofir Nachum, Carolina Parada, Jodilyn Peralta, Emily Perez, Karl
  Pertsch, Jornell Quiambao, Kanishka Rao, Michael Ryoo, Grecia Salazar, Pannag
  Sanketi, Kevin Sayed, Jaspiar Singh, Sumedh Sontakke, Austin Stone, Clayton
  Tan, Huong Tran, Vincent Vanhoucke, Steve Vega, Quan Vuong, Fei Xia, Ted
  Xiao, Peng Xu, Sichun Xu, Tianhe Yu, and Brianna Zitkovich.
\newblock Rt-1: Robotics transformer for real-world control at scale.
\newblock In {\em arXiv preprint arXiv:2212.06817}, 2022.

\bibitem{mbm2021}
Constantinos Chamzas, Carlos Quintero-Pena, Zachary Kingston, Andreas Orthey,
  Daniel Rakita, Michael Gleicher, Marc Toussaint, and Lydia~E. Kavraki.
\newblock Motionbenchmaker: A tool to generate and benchmark motion planning
  datasets.
\newblock In {\em IEEE Robotics and Automation Letters}. IEEE, 2021.

\bibitem{shapenet}
A.X. Chang, T.~Funkhouser, L.~Guibas, Pat Hanrahan, Q.~Huang, Z~Li,
  S.~Savarese, M.~Savva, S.~Song, H.~Su, J.~Xiao, L.~Yi, and F.~Yu.
\newblock Shapenet: An information-rich 3d model repository.
\newblock {\em Technical report, Stanford University — Princeton University
  — Toyota Technological Institute at Chicago}, 2015.

\bibitem{pointnet}
Hao~Su Charles R~Qi, Li~Yi and Leonidas~J Guibas.
\newblock Pointnet++: Deep hierarchical feature learning on point sets in a
  metric space.
\newblock {\em Neural Information Processing Systems (NeurIPS)}, 2017.

\bibitem{chitta2012moveit}
Sachin Chitta, Ioan Sucan, and Steve Cousins.
\newblock Moveit![ros topics].
\newblock {\em IEEE Robotics \& Automation Magazine}, 19(1):18--19, 2012.

\bibitem{danielczuk2019}
Michael Danielczuk, Matthew Matl, Saurabh Gupta, Andrew Li, Andrew Lee, Jeff
  Mahler, and Ken Goldberg.
\newblock Segmenting unknown 3d objects from real depth images using mask r-cnn
  trained on synthetic data.
\newblock In {\em IEEE International Conference on Robotics and Automation
  (ICRA)}. IEEE, 2019.

\bibitem{Danielczuk2021ObjectRU}
Michael Danielczuk, Arsalan Mousavian, Clemens Eppner, and Dieter Fox.
\newblock Object rearrangement using learned implicit collision functions.
\newblock {\em 2021 IEEE International Conference on Robotics and Automation
  (ICRA)}, pages 6010--6017, 2021.

\bibitem{yip2020}
Nikhil Das and Michael Yip.
\newblock Learning-based proxy collision detection for robot motion planning
  applications.
\newblock {\em Transactions on Robotics}, 2021.

\bibitem{deitke2022procthor}
Matt Deitke, Eli VanderBilt, Alvaro Herrasti, Luca Weihs, Jordi Salvador, Kiana
  Ehsani, Winson Han, Eric Kolve, Ali Farhadi, Aniruddha Kembhavi, et~al.
\newblock Procthor: Large-scale embodied ai using procedural generation.
\newblock {\em arXiv preprint arXiv:2206.06994}, 2022.

\bibitem{VisionTransformers2020}
Alexey Dosovitskiy, Lucas Beyer, Alexander Kolesnikov, Dirk Weissenborn,
  Xiaohua Zhai, Thomas Unterthiner, Mostafa Dehghani, Matthias Minderer, Georg
  Heigold, Sylvain Gelly, and Jakob~Uszkoreit andNeil Houlsby.
\newblock Attention is all you need.
\newblock In {\em In International Conference on Learning Representations},
  2021.

\bibitem{ehsani2021manipulathor}
Kiana Ehsani, Winson Han, Alvaro Herrasti, Eli VanderBilt, Luca Weihs, Eric
  Kolve, Aniruddha Kembhavi, and Roozbeh Mottaghi.
\newblock Manipulathor: A framework for visual object manipulation.
\newblock In {\em The IEEE Conference on Computer Vision and Pattern
  Recognition (CVPR)}, 2021.

\bibitem{acronym2020}
Clemens Eppner, Arsalan Mousavian, and Dieter Fox.
\newblock {ACRONYM}: A large-scale grasp dataset based on simulation.
\newblock In {\em 2021 {IEEE} Int. Conf. on Robotics and Automation, {ICRA}},
  2020.

\bibitem{fisher2012example}
Matthew Fisher, Daniel Ritchie, Manolis Savva, Thomas Funkhouser, and Pat
  Hanrahan.
\newblock Example-based synthesis of 3d object arrangements.
\newblock {\em ACM Transactions on Graphics (TOG)}, 31(6):1--11, 2012.

\bibitem{fishman2022mpinets}
Adam Fishman, Adithyavairavan Murali, Clemens Eppner, Bryan Peele, Byron Boots,
  and Dieter Fox.
\newblock Motion policy networks.
\newblock {\em Conference on Robot Learning (CoRL)}, 2022.

\bibitem{fu20213d}
Huan Fu, Bowen Cai, Lin Gao, Ling-Xiao Zhang, Jiaming Wang, Cao Li, Qixun Zeng,
  Chengyue Sun, Rongfei Jia, Binqiang Zhao, et~al.
\newblock 3d-front: 3d furnished rooms with layouts and semantics.
\newblock In {\em Proceedings of the IEEE/CVF International Conference on
  Computer Vision}, pages 10933--10942, 2021.

\bibitem{KaolinLibrary}
Clement Fuji~Tsang, Maria Shugrina, Jean~Francois Lafleche, Towaki Takikawa,
  Jiehan Wang, Charles Loop, Wenzheng Chen, Krishna~Murthy Jatavallabhula,
  Edward Smith, Artem Rozantsev, Or~Perel, Tianchang Shen, Jun Gao, Sanja
  Fidler, Gavriel State, Jason Gorski, Tommy Xiang, Jianing Li, Michael Li, and
  Rev Lebaredian.
\newblock Kaolin: A pytorch library for accelerating 3d deep learning research.
\newblock \url{https://github.com/NVIDIAGameWorks/kaolin}, 2022.

\bibitem{mppi2017}
A.~Aldrich G.~Williams and E.~A. Theodorou.
\newblock Model predictive path integral control: From theory to parallel
  computation.
\newblock {\em Journal of Guidance, Control, and Dynamics}, 2017.

\bibitem{Gammell2015BatchIT}
Jonathan~D. Gammell, Siddhartha~S. Srinivasa, and Tim~D. Barfoot.
\newblock Batch informed trees (bit*): Sampling-based optimal planning via the
  heuristically guided search of implicit random geometric graphs.
\newblock {\em 2015 IEEE International Conference on Robotics and Automation
  (ICRA)}, pages 3067--3074, 2015.

\bibitem{garrett2021}
Caelan~Reed Garrett, Rohan Chitnis, Rachel Holladay, Beomjoon Kim, Tom Silver,
  Leslie~Pack Kaelbling, and Tomás Lozano-Pérez.
\newblock Integrated task and motion planning.
\newblock {\em Annual Review of Control, Robotics, and Autonomous Systems},
  2021.

\bibitem{garrett2020onlinereplanning}
Caelan~Reed Garrett, Chris Paxton, Tomás Lozano-Pérez, Leslie~Pack Kaelbling,
  and Dieter Fox.
\newblock Online replanning in belief space for partially observable task and
  motion problems.
\newblock In {\em IEEE International Conference on Robotics and Automation
  (ICRA)}. IEEE, 2020.

\bibitem{gan2014}
Ian Goodfellow, Jean Pouget-Abadie, Mehdi Mirza, Bing Xu, David Warde-Farley,
  Sherjil Ozair, Aaron Courville, and Yoshua Bengio.
\newblock Generative adversarial nets.
\newblock {\em Neural Information Processing Systems}, 2014.

\bibitem{goodwin2022semantically}
Walter Goodwin, Sagar Vaze, Ioannis Havoutis, and Ingmar Posner.
\newblock Semantically grounded object matching for robust robotic scene
  rearrangement.
\newblock In {\em 2022 International Conference on Robotics and Automation
  (ICRA)}, pages 11138--11144. IEEE, 2022.

\bibitem{ifor2022}
Ankit Goyal, Arsalan Mousavian, Chris Paxton, Yu-Wei Chao, Brian Okorn, Jia
  Deng, and Dieter Fox.
\newblock Ifor: Iterative flow minimization for robotic object rearrangement.
\newblock {\em Proceedings of the IEEE/CVF Conference on Computer Vision and
  Pattern Recognition (CVPR)}, 2022.

\bibitem{robotsinhomenips2018}
Abhinav Gupta, Adithyavairavan Murali, Dhiraj Gandhi, and Lerrel Pinto.
\newblock Robot learning in homes: Improving generalization and reducing
  dataset bias.
\newblock {\em Neural Information Processing Systems (NeurIPS)}, 2018.

\bibitem{Hauser2010FastSO}
Kris~K. Hauser and Victor Ng-Thow-Hing.
\newblock Fast smoothing of manipulator trajectories using optimal
  bounded-acceleration shortcuts.
\newblock {\em 2010 IEEE International Conference on Robotics and Automation},
  pages 2493--2498, 2010.

\bibitem{nistenv2021}
John Horst, Jeremy Marvel, and Elena Messina.
\newblock Best practices for the integration of collaborative robots into
  workcells within small and medium-sized manufacturing operations.
\newblock {\em NIST Advanced Manufacturing Series}, 100(41), 2021.

\bibitem{huang2022equivariant}
Haojie Huang, Dian Wang, Robin Walter, and Robert Platt.
\newblock Equivariant transporter network.
\newblock {\em arXiv preprint arXiv:2202.09400}, 2022.

\bibitem{hubbard1996approximating}
Philip~M Hubbard.
\newblock Approximating polyhedra with spheres for time-critical collision
  detection.
\newblock {\em ACM Transactions on Graphics (TOG)}, 15(3):179--210, 1996.

\bibitem{james2019rlbench}
Stephen James, Zicong Ma, David Rovick~Arrojo, and Andrew~J. Davison.
\newblock Rlbench: The robot learning benchmark \& learning environment.
\newblock {\em IEEE Robotics and Automation Letters}, 2020.

\bibitem{kar2019metasim}
Amlan Kar, Aayush Prakash, Ming-Yu Liu, Eric Cameracci, Justin Yuan, Matt
  Rusiniak, David Acuna, Antonio Torralba, and Sanja Fidler.
\newblock Meta-sim: Learning to generate synthetic datasets.
\newblock In {\em International Conference on Computer Vision (ICCV)}, 2019.

\bibitem{faust2021}
J.~Chase Kew, Brian Ichter, Maryam Bandari, Tsang-Wei~Edward Lee, and
  Aleksandra Faust.
\newblock Neural collision clearance estimator for batched motion planning.
\newblock 2021.

\bibitem{klingensmith2016}
Matthew Klingensmith, Siddartha Sirinivasa, and Michael Kaess.
\newblock Articulated robot motion for simultaneous localization and mapping
  (arm-slam).
\newblock In {\em Robotics and Automation Letters}. IEEE, 2016.

\bibitem{rrtconnect}
James Kuffner and Steven~M. LaValle.
\newblock Rrt-connect: An efficient approach to single-query path planning.
\newblock In {\em International Conference on Robotics and Automation (ICRA)}.
  IEEE, 2000.

\bibitem{labbe2020monte}
Yann Labb{\'e}, Sergey Zagoruyko, Igor Kalevatykh, Ivan Laptev, Justin
  Carpentier, Mathieu Aubry, and Josef Sivic.
\newblock Monte-carlo tree search for efficient visually guided rearrangement
  planning.
\newblock {\em IEEE Robotics and Automation Letters}, 5(2):3715--3722, 2020.

\bibitem{li2021igibson}
Chengshu Li, Fei Xia, Roberto Mart{\'\i}n-Mart{\'\i}n, Michael Lingelbach,
  Sanjana Srivastava, Bokui Shen, Kent~Elliott Vainio, Cem Gokmen, Gokul
  Dharan, Tanish Jain, Andrey Kurenkov, Karen Liu, Hyowon Gweon, Jiajun Wu,
  Li~Fei-Fei, and Silvio Savarese.
\newblock igibson 2.0: Object-centric simulation for robot learning of everyday
  household tasks.
\newblock In {\em Conference on Robot Learning}, 2021.

\bibitem{imle2018}
Ke~Li and Jitendra Malik.
\newblock Implicit maximum likelihood estimation.
\newblock {\em arXiv preprint arXiv:1809.09087}, 2018.

\bibitem{mahler2018binpicking}
Jeffrey Mahler and Ken Goldberg.
\newblock Learning deep policies for robot bin picking by simulating robust
  grasping sequences.
\newblock {\em Conference on Robot Learning}, 2017.

\bibitem{mahler2017dex}
Jeffrey Mahler, Jacky Liang, Sherdil Niyaz, Michael Laskey, Richard Doan, Xinyu
  Liu, Juan~Aparicio Ojea, and Ken Goldberg.
\newblock Dex-net 2.0: Deep learning to plan robust grasps with synthetic point
  clouds and analytic grasp metrics.
\newblock {\em RSS}, 2017.

\bibitem{majerowicz2013filling}
Lucas Majerowicz, Ariel Shamir, Alla Sheffer, and Holger~H Hoos.
\newblock Filling your shelves: Synthesizing diverse style-preserving artifact
  arrangements.
\newblock {\em IEEE transactions on visualization and computer graphics},
  20(11):1507--1518, 2013.

\bibitem{Mildenhall2020NeRFRS}
Ben Mildenhall, Pratul~P. Srinivasan, Matthew Tancik, Jonathan~T. Barron, Ravi
  Ramamoorthi, and Ren Ng.
\newblock Nerf: Representing scenes as neural radiance fields for view
  synthesis.
\newblock In {\em European Conference on Computer Vision (ECCV)}, 2020.

\bibitem{Mo_2019_CVPR}
Kaichun Mo, Shilin Zhu, Angel~X. Chang, Li~Yi, Subarna Tripathi, Leonidas~J.
  Guibas, and Hao Su.
\newblock {PartNet}: A large-scale benchmark for fine-grained and hierarchical
  part-level {3D} object understanding.
\newblock In {\em The IEEE Conference on Computer Vision and Pattern
  Recognition (CVPR)}, June 2019.

\bibitem{6dofgraspnet}
Arsalan Mousavian, Clemens Eppner, and Dieter Fox.
\newblock {6-DOF} {GraspNet}: Variational grasp generation for object
  manipulation.
\newblock {\em International Conference on Computer Vision}, 2019.

\bibitem{objseeker}
Arsalan Mousavian, Lucas Manuelli, Brian Okorn, Yu~Xiang, Clemens Eppner,
  Adithyavairavan Murali, and Dieter Fox.
\newblock Objectseeker: A unified framework for one-shot object detection,
  tracking, and instance segmentation of everyday objects.
\newblock In {\em arXiv}, 2023.

\bibitem{Murali2020CollisionNet}
Adithyavairavan Murali, Arsalan Mousavian, Clemens Eppner, Chris Paxton, and
  Dieter Fox.
\newblock 6-dof grasping for target-driven object manipulation in clutter.
\newblock In {\em IEEE International Conference on Robotics and Automation
  (ICRA)}, 2020.

\bibitem{KinectFusion2011}
Richard Newcombe, Shahram Izadi, Otmar Hilliges, David Molyneaux, David Kim,
  Andrew~J. Davison, Pushmeet Kohli, Jamie Shotton, Steve Hodges, and Andrew
  Fitzgibbon.
\newblock Kinectfusion: Real-time dense surface mapping and tracking.
\newblock {\em 2011 10th IEEE International Symposium on Mixed and Augmented
  Reality (ISMAR)}, 2011.

\bibitem{fcl2012}
Jia Pan, Sachin Chitta, and Dinesh Manocha.
\newblock Fcl: A general purpose library for collision and proximity queries.
\newblock {\em 2012 IEEE International Conference on Robotics and Automation
  (ICRA)}, 2012.

\bibitem{nerp2021}
Ahmed~H Qureshi, Arsalan Mousavian, Chris Paxton, Michael Yip, and Dieter Fox.
\newblock Nerp: Neural rearrangement planning for unknown objects.
\newblock {\em RSS}, 2021.

\bibitem{CLIP2021}
Alec Radford, Jong~Wook Kim, Chris Hallacy, Aditya Ramesh, Gabriel Goh,
  Sandhini Agarwal, Girish Sastry, Amanda Askell, Pamela Mishkin, Jack Clark,
  Gretchen Krueger, and Ilya Sutskever.
\newblock Learning transferable visual models from natural language
  supervision.
\newblock In {\em arXiv:2103.00020}, 2021.

\bibitem{tamp204}
Siddharth Srivastava, Eugene Fang, Lorenzo Riano, Rohan Chitnis, Stuart
  Russell, and Pieter Abbeel.
\newblock Combined task and motion planning through an extensible
  planner-independent interface layer.
\newblock In {\em IEEE International Conference on Robotics and Automation
  (ICRA)}, 2014.

\bibitem{replica19arxiv}
Julian Straub, Thomas Whelan, Lingni Ma, Yufan Chen, Erik Wijmans, Simon Green,
  Jakob~J. Engel, Raul Mur-Artal, Carl Ren, Shobhit Verma, Anton Clarkson,
  Mingfei Yan, Brian Budge, Yajie Yan, Xiaqing Pan, June Yon, Yuyang Zou,
  Kimberly Leon, Nigel Carter, Jesus Briales, Tyler Gillingham, Elias Mueggler,
  Luis Pesqueira, Manolis Savva, Dhruv Batra, Hauke~M. Strasdat, Renzo~De
  Nardi, Michael Goesele, Steven Lovegrove, and Richard Newcombe.
\newblock The {R}eplica dataset: A digital replica of indoor spaces.
\newblock {\em arXiv preprint arXiv:1906.05797}, 2019.

\bibitem{sundermeyer2021contact}
Martin Sundermeyer, Arsalan Mousavian, Rudolph Triebel, and Dieter Fox.
\newblock Contact-graspnet: Efficient 6-dof grasp generation in cluttered
  scenes.
\newblock 2021.

\bibitem{szot2021habitat}
Andrew Szot, Alex Clegg, Eric Undersander, Erik Wijmans, Yili Zhao, John
  Turner, Noah Maestre, Mustafa Mukadam, Devendra Chaplot, Oleksandr Maksymets,
  Aaron Gokaslan, Vladimir Vondrus, Sameer Dharur, Franziska Meier, Wojciech
  Galuba, Angel Chang, Zsolt Kira, Vladlen Koltun, Jitendra Malik, Manolis
  Savva, and Dhruv Batra.
\newblock Habitat 2.0: Training home assistants to rearrange their habitat.
\newblock In {\em Advances in Neural Information Processing Systems (NeurIPS)},
  2021.

\bibitem{9682604}
Karl Van~Wyk, Mandy Xie, Anqi Li, Muhammad~Asif Rana, Buck Babich, Bryan Peele,
  Qian Wan, Iretiayo Akinola, Balakumar Sundaralingam, Dieter Fox, Byron Boots,
  and Nathan~D. Ratliff.
\newblock Geometric fabrics: Generalizing classical mechanics to capture the
  physics of behavior.
\newblock {\em IEEE Robotics and Automation Letters}, 7(2):3202--3209, 2022.

\bibitem{issacgym2021}
Yunrong Guo Michelle Lu Kier Storey Miles Macklin David Hoeller Nikita Rudin
  Arthur Allshire Ankur Handa Gavriel~State Viktor~Makoviychuk,
  Lukasz~Wawrzyniak.
\newblock Isaac gym: High performance gpu-based physics simulation for robot
  learning.
\newblock {\em arXiv:2108.10470}, 2021.

\bibitem{Wada:etal:ICRA2022b}
Kentaro Wada, Stephen James, and Andrew~J. Davison.
\newblock {ReorientBot}: Learning object reorientation for specific-posed
  placement.
\newblock In {\em IEEE International Conference on Robotics and Automation
  (ICRA)}, 2022.

\bibitem{wu2022transporters}
Hongtao Wu, Jikai Ye, Xin Meng, Chris Paxton, and Gregory Chirikjian.
\newblock Transporters with visual foresight for solving unseen rearrangement
  tasks.
\newblock {\em arXiv preprint arXiv:2202.10765}, 2022.

\bibitem{Xiang_2020_SAPIEN}
Fanbo Xiang, Yuzhe Qin, Kaichun Mo, Yikuan Xia, Hao Zhu, Fangchen Liu, Minghua
  Liu, Hanxiao Jiang, Yifu Yuan, He~Wang, Li~Yi, Angel~X. Chang, Leonidas~J.
  Guibas, and Hao Su.
\newblock {SAPIEN}: A simulated part-based interactive environment.
\newblock In {\em The IEEE Conference on Computer Vision and Pattern
  Recognition (CVPR)}, June 2020.

\bibitem{yu2015clutterpalette}
Lap-Fai Yu, Sai-Kit Yeung, and Demetri Terzopoulos.
\newblock The clutterpalette: An interactive tool for detailing indoor scenes.
\newblock {\em IEEE transactions on visualization and computer graphics},
  22(2):1138--1148, 2015.

\bibitem{yu2019meta}
Tianhe Yu, Deirdre Quillen, Zhanpeng He, Ryan Julian, Karol Hausman, Chelsea
  Finn, and Sergey Levine.
\newblock Meta-world: A benchmark and evaluation for multi-task and meta
  reinforcement learning.
\newblock In {\em Conference on Robot Learning (CoRL)}, 2019.

\bibitem{zeng2020transporter}
Andy Zeng, Pete Florence, Jonathan Tompson, Stefan Welker, Jonathan Chien,
  Maria Attarian, Travis Armstrong, Ivan Krasin, Dan Duong, Vikas Sindhwani,
  et~al.
\newblock Transporter networks: Rearranging the visual world for robotic
  manipulation.
\newblock {\em Conference on Robot Learning}, 2020.

\bibitem{Zhou2018}
Qian-Yi Zhou, Jaesik Park, and Vladlen Koltun.
\newblock {Open3D}: {A} modern library for {3D} data processing.
\newblock {\em arXiv:1801.09847}, 2018.

\end{thebibliography}
